\documentclass[runningheads]{llncs}
\usepackage{graphicx}

\usepackage{tikz}
\usepackage{comment} 
\usepackage{amsmath,amssymb} 
\usepackage{color}

\usepackage{multirow}
\usepackage{ulem}

\usepackage{subcaption}
\usepackage{bm}
\usepackage{amsmath}
\usepackage{adjustbox}
\usepackage[colorlinks = true,
            linkcolor = red,
            urlcolor  = green,
            citecolor = black,
            anchorcolor = red]{hyperref}

\newcommand{\etal}{\textit{et al. }}

\begin{document}
\pagestyle{headings}
\mainmatter
\def\ECCVSubNumber{4125}  

\title{Kinship Identification through Joint Learning using Kinship Verification Ensembles
} 

\titlerunning{ECCV-20 submission ID \ECCVSubNumber} 
\authorrunning{ECCV-20 submission ID \ECCVSubNumber} 
\author{Anonymous ECCV submission}
\institute{Paper ID \ECCVSubNumber}

\titlerunning{Kinship Identification through Joint Learning}
%
\author{Wei Wang\inst{1},
Shaodi You\inst{1}, 
Sezer Karaoglu\inst{2},
Theo Gevers\inst{1,2}}
\authorrunning{W. Wang et al.}
%
\institute{University of Amsterdam  \\
\email{\{w.wang,s.you,th.gevers\}@uva.nl}\and
3DUniversum  \\
\email{s.karaoglu@3duniversum.com}}
\maketitle

\begin{abstract}
Kinship verification is a well-explored task: identifying whether or not two persons are kin. In contrast, kinship identification has been largely ignored so far. Kinship identification aims to further identify the particular type of kinship. An extension to kinship verification run short to properly obtain identification, because existing verification networks are individually trained on specific kinships and do not consider the context between different kinship types. Also, existing kinship verification datasets have biased positive-negative distributions which are different than real-world distributions. 

To this end, we propose a novel kinship identification approach based on joint training of kinship verification ensembles and classification modules. We propose to rebalance the training dataset to become more realistic. Large scale experiments demonstrate the appealing performance on kinship identification. The experiments further show significant performance improvement of kinship verification when trained on the same dataset with more realistic distributions.
\keywords{kinship identification, kinship verification ensemble, joint learning}
\end{abstract}
  
\section{Introduction}

Kinship is the relationship between people who are biologically related with overlapping genes \cite{kinwild,kinwild2}, such as parent-children, sibling-sibling, and grandparent-grandchildren \cite{almuashi2017automated,FIW,rfiw,latentadaptive}. Image-based kinship identification is used in a variety of applications including missing children searching \cite{latentadaptive}, family album organization, forensic investigation \cite{rfiw}, automatic image annotation \cite{kinwild}, social media analysis \cite{social1,social3,social4}, social behavior analysis \cite{social2,kindeep,social6,social7}, historical and genealogical research \cite{social5,social3}, and crime scene investigation \cite{MixedAutoencoder}. 

\begin{figure}[!t]
    \centering
    \includegraphics[width=1\textwidth]{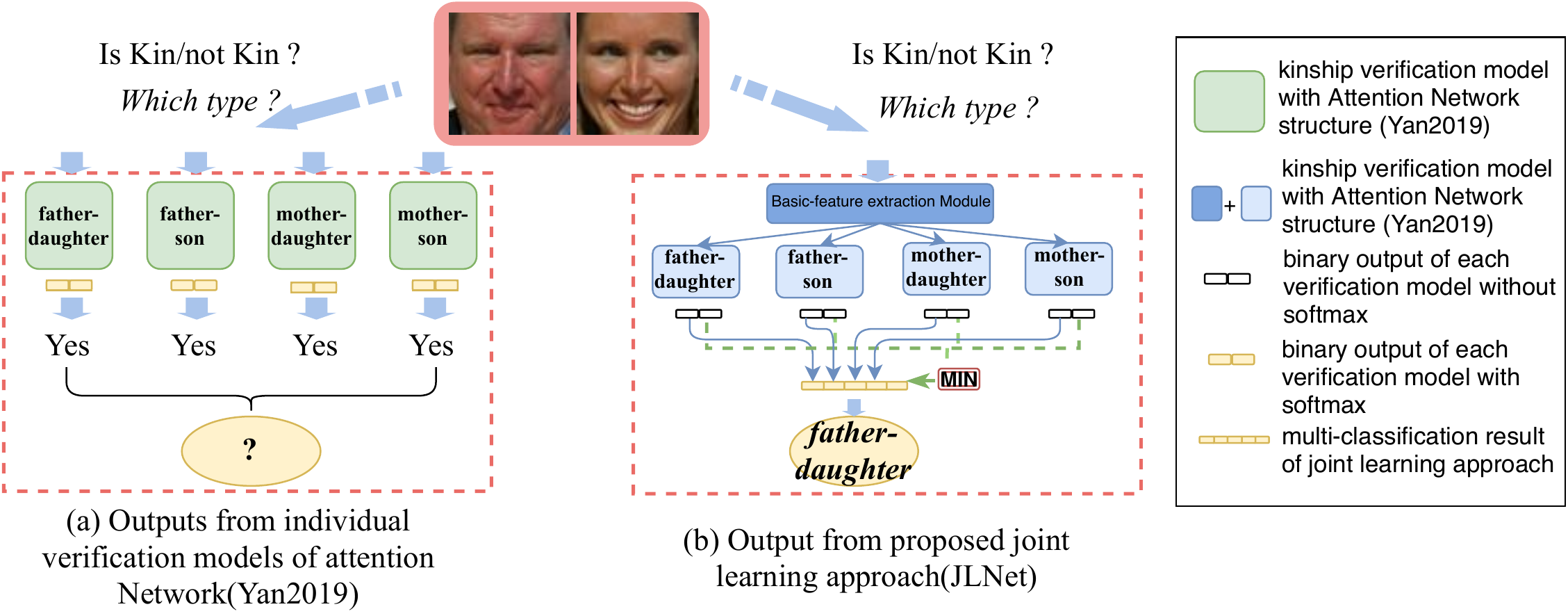}
    \caption{Identification of kinship relationships using verification ensembles.
    (a) Existing verification networks are trained independently resulting in contradictory outputs.
    (b) The output of our proposed joint training
    }
    \label{fig:intro_example}
\end{figure}

While kinship verification is a well-explored task, identifying whether or not persons are kin, kinship identification, which is the task to further identify the particular type of kinship, has been largely ignored so far. 
Existing kinship verification methods usually train and test each type of kinship model independently \cite{wang2018cross,FIW,latentadaptive} and hence do not
fully exploit the complementary information among different kin types. 
Moreover, existing datasets have unrealistic positive-negative sample distributions. This leads to significant limitations in real world applications. When conducting kinship identification, since there is no prior knowledge of the
distribution of images, all independently trained models are used to determine the kinship type of a specific image pair. Fig.~\ref{fig:intro_example} shows an example of providing an image pair to four individually trained verification networks based on a recent state-of-the-art method by Yan \etal \cite{attenNet}. The network generates contradictory outputs showing that the test subjects are simultaneously father-daughter, father-son, mother-son and mother-daughter.

In this paper, a new identification method is proposed to learn the identification and verification labels jointly i.e. combining the kinship identification and verification tasks. Specifically, all kinship-type verification models are ensembled by combining the binary output of each verification model to form a multi-class output while training. The binary and multi-class models are leveraged in a multi-task-learning way during the training process to enhance generalization capabilities. Also, we propose a baseline multi-classification neural network for comparison.

We test our proposed kinship identification method on the KinfaceWI and KinfaceWII datasets and demonstrate state-of-the-art performance for kinship identification. We also show that the proposed method significantly improves the performance of kinship verification when trained on the same unbiased dataset.

To summarize, the contributions of our work are:
\begin{itemize}
    \item We propose a theoretical analysis in metric space of relationships between kinship identification and kinship verification.
    \item We propose a joint learnt network that simultaneously optimizes the performance of kinship verification and kinship identification.
    \item The proposed method outperforms existing methods for both kinship identification and unbiased kinship verification.
\end{itemize}
\section{Related Work}
\paragraph{Kinship Verification}
Fang \etal \cite{first} are the first to use handcrafted feature descriptors for kinship verification. Later, Xia \etal collected a new dataset with young and old parent images to utilize the intermediate distribution using transfer learning \cite{xia2011kinship,transferlearning2}. Lu \etal \cite{kinwild2,zhou2011kinship} propose a series of metric learning methods. Other handcrafted feature-based methods can be found in \cite{yan2017facial,wang2009hog,xia2011kinship,zhou2012gabor,hamdi1,kinwild,DMML,family101}. Deep learning-based methods \cite{kindeep,attenNet} exploits the advantages of deep feature representations by using pre-trained neural networks in an off-the-shelf way. Zhang \etal are the first to use deep convolutional neural networks \cite{kindeep}, and Yan \etal \cite{attenNet} are the first to add attention mechanisms in deep learning networks for kinship verification. In recent years, there is a trend to combine different features from both traditional descriptors \cite{yan2017facial,zhou2011kinship} and deep neural networks \cite{deepcite1,deepcite2,deepcite3} to generate better representations \cite{comb2}. (m)DML \cite{DML,mDML} combines auto-encoders with metric learning. However, these methods focus on specific types of kinship and train and test on the same kinship types separately, which may not be feasible in real-world scenarios.
\paragraph{Kinship Identification} Different from kinship verification, kinship identification attracted less attention \cite{almuashi2017automated}. \cite{almuashi2017automated,FIW} only slightly deal with kinship identification. Guo \etal \cite{graphbased} propose a pairwise kinship identification method using a multi-class linear logistic regressor. The method uses graph information from one image with multi inputs. The paper is based on "kinship recognition" and uses a strong assumption that all the data is processed by a perfect kinship verification algorithm. Since there is not sufficient data with family annotations, the method is limited by using multi-input labels. In contrast, our method handles negative pairs and focuses on pair-wise kinship identification. For example, in the context of searching for missing children, we need to handle each potential pair online and find the most likely pair for specific kinship types. In this case, we need to filter the online data and test the most likely data after filtering. As for the family photo arrangement or social media analysis, the aim is to understand the relationships between persons in a picture. There are usually many faces and different kinship relations in a family picture. Hence, the goal is to verify the most likely pairs among negative pairs. Previous methods are not able to cope with this scenario.
Fig \ref{fig:KR_relation_structure} shows that kinship verification is closely related to kinship identification. As a consequence, we propose a new approach by jointly learning all independent models with kinship verification and identification information. 

\section{Kinship Identification through Joint Learning with Kinship Verification}
In this section, we first introduce the three types of relationship understanding: kinship verification, kinship identification, and kinship classification. Based on this, we introduce the current challenge on kinship identification. Finally, we introduce the concept of conducting kinship identification by using a joint learning strategy between kinship identification and kinship verification.
\subsection{Definition of Kinship Verification, Kinship Identification and Kinship Classification}
\begin{figure}[tb]
    \centering
        \includegraphics[width=1\textwidth]{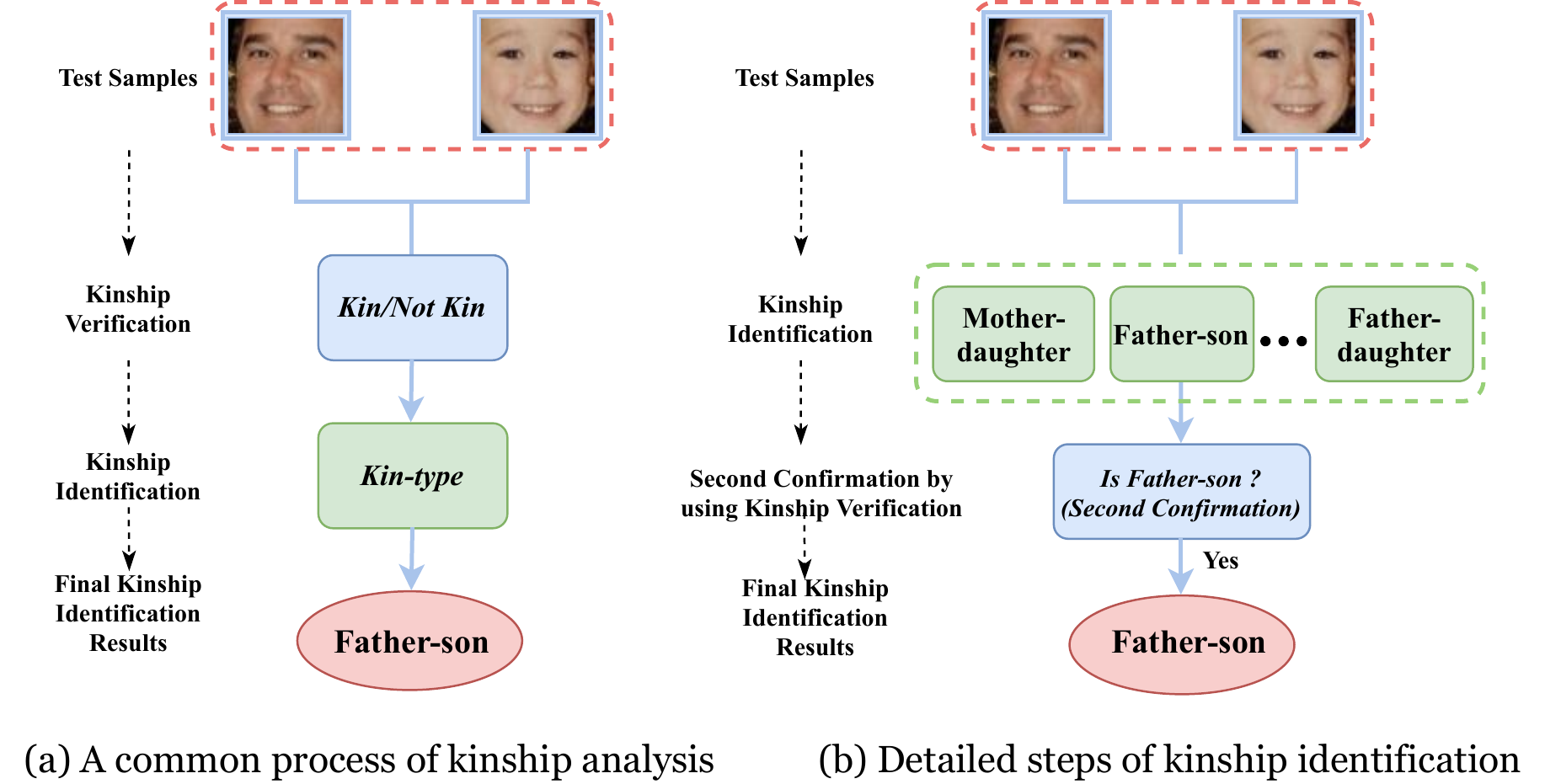}
    \caption{Flowchart of the relation between kinship verification and kinship identification. (a) Kinship  verification  is  used  as  a  preliminary  process  for  kinship  identification. (b) The kinship identification process can be divided into two steps: kinship identification and kinship verification on a specific type.}
    \label{fig:KR_relation_structure}
\end{figure}
Kinship recognition is the general task of kinship analysis based on visual information. There are mainly three sub-tasks \cite{FIW,almuashi2017automated}: kinship verification, kinship identification, and kinship classification (e.g. family recognition). The goal of kinship verification is to authenticate the relationship between image pairs of persons by determining whether they are blood-related or not. Kinship identification aims at determining the type of kinship relation between persons. Kinship classification \cite{latentadaptive,FIW} is the recognition of the family to which a person belongs to. Fig.~\ref{fig:KR_relation_structure} illustrates the relationship between these tasks. This paper focuses on kinship identification, which is an important but not well-explored topic. Unlike other kinship recognition methods \cite{wang2017leveraging,chen2012discovering,wang2018photo,graphbased}, which take images of multiple people as input to predict the relationships between them, the kinship identification task targets at classifying the kin-type of image pairs (negative pairs also included).
    
\subsection{Relationship between Kinship Verification and Kinship Identification and the Limitation of Existing Methods}

\subsubsection{Relation between the Two Tasks}
In the literature, kinship verification and identification are two tasks which are studied separately but are closely related. When analyzing the kinship relation between persons, verification is usually applied first to determine whether these persons are kin or not. Then, the kinship type is defined. Fig.~\ref{fig:KR_relation_structure}.a shows the common process of kinship analysis, where kinship verification is used as a preliminary process for kinship identification. Furthermore, kinship identification can be divided into two steps, as shown in Fig.~\ref{fig:KR_relation_structure}.b. In the first step, the images are preliminarily classified by the kinship identification model. Then, the classified images are sent to the corresponding verification model. Due to the differences of inherited features among different kin-type images, the kinship verification model provides a better representation than a general kinship identification model. On the other hand, since the kinship identification process filters out irrelevant samples, it provides a consistent and similar feature distribution for kinship verification modelling. In this way, kinship verification and identification are two complementary processes, and can benefit from each other.  
\begin{figure}[tbp]

    \centering
        
        \begin{subfigure}[t]{0.31\textwidth}
            \includegraphics[width=\textwidth]{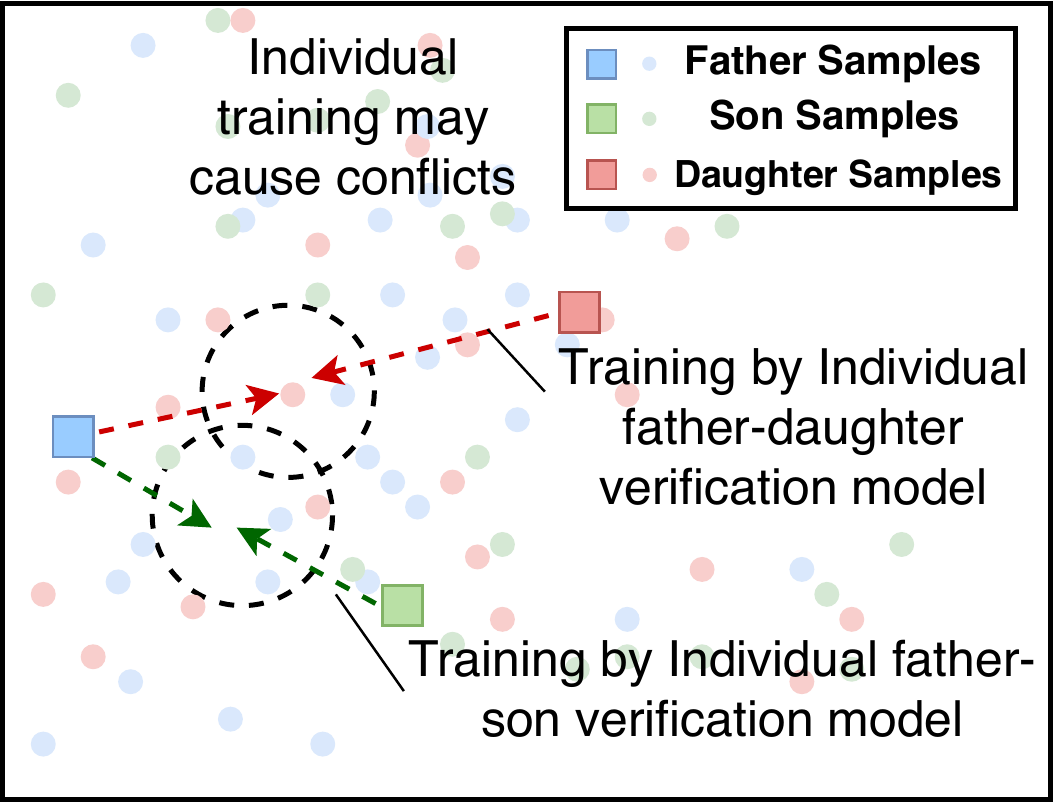}
            \subcaption{Feature learning phase of different individual trained models for the kinship verification task}
            \label{fig3:ml1}
        \end{subfigure}
        ~
        \begin{subfigure}[t]{0.31\textwidth}
            \includegraphics[width=\textwidth]{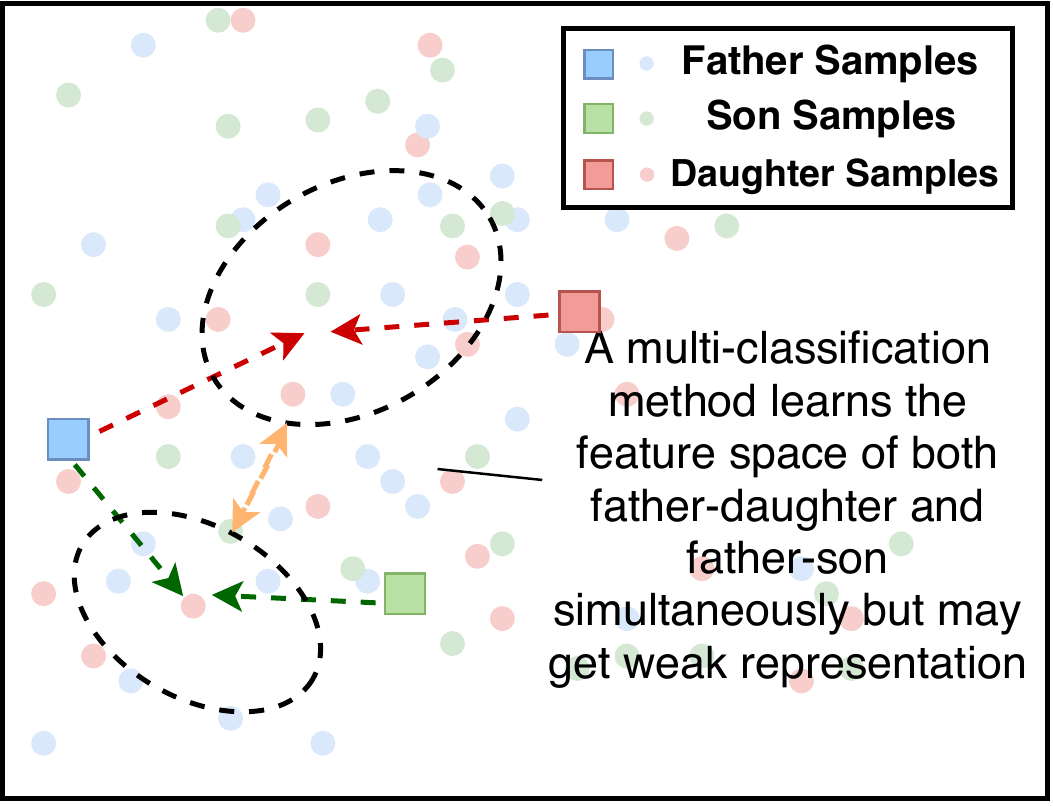}
            \subcaption{Feature learning phase of a multi-classification method for kinship identification task}
            \label{fig3:ml2}
        \end{subfigure}
        ~
        \begin{subfigure}[t]{0.31\textwidth}
            \includegraphics[width=\textwidth]{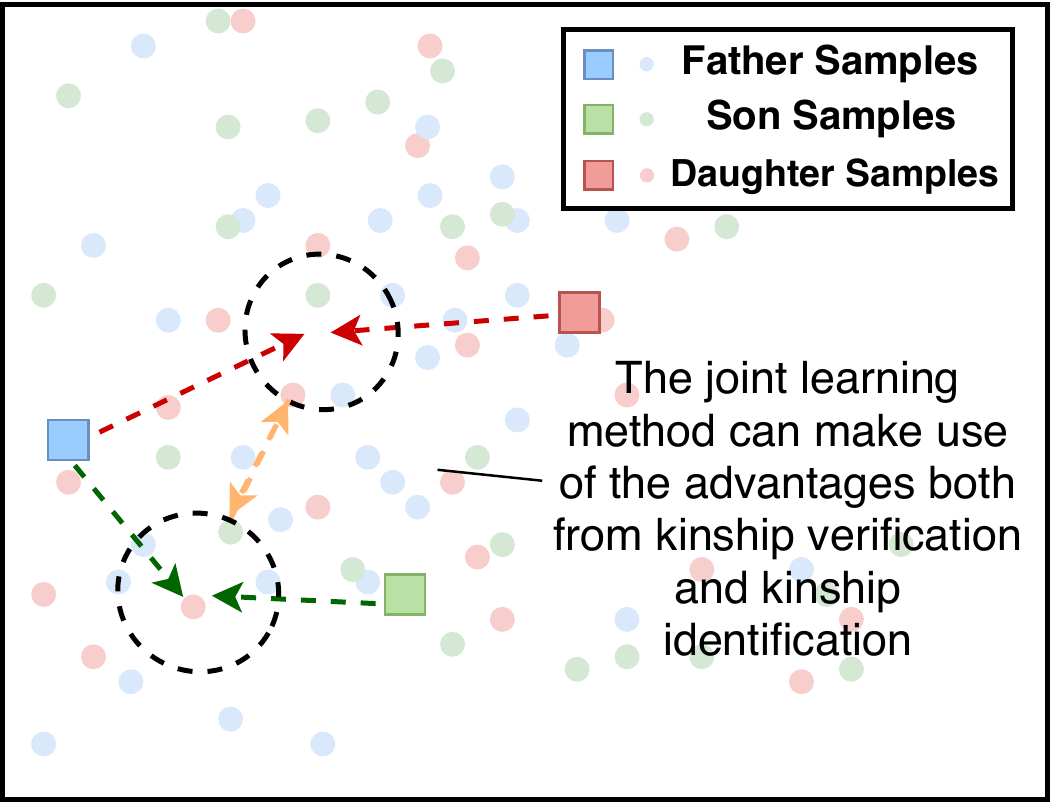}
            \subcaption{Feature learning phase of joint learning by using kinship identification and verification}
            \label{fig3:ml3}
        \end{subfigure}
        
    \caption{Feature space of models during training. Similar feature shapes indicate that the samples are from the same family. Joint learning better represents the context between different kinship relationships. Small circles are used to represent focused samples in feature space.}
    \label{fig:approach_alg_independent_learning}
\end{figure} 
\subsubsection{Representation of Kinship Relationships in Metric Feature Space and Limitation of Existing Methods}
In the literature, metric learning is a popular approach for kinship verification. Ideally, the learnt metric space represents kinship likeness for smaller distances. However, existing kinship verification models only consider specific kinship types and ignore the influence of other types. 

As shown in Fig.~\ref{fig3:ml1}, when the father-daughter verification model is being trained, the features of father and daughter samples will be congregated  during the training process and the negative daughter images will be pulled apart. However, due to the negative samples of father-son pairs, which are not included in the training data, the features of son images are less affected by the training process pulling father-son images apart. A narrow-down training of kinship verification can improve the representation of each sample within a specific kin-type. However, since the model does not thoroughly learn other types of negative samples, the separate trained models can easily conflict with each other resulting in ambiguous results. In contrast, a multi-classification method not only considers different types of images but also the interaction between different types. As shown in Fig.~\ref{fig3:ml2}, the son features will be learned as negative features for the father-daughter feature space, whereas the features of daughters will be considered as negative features for the father-son space. The yellow arrows in Fig.~\ref{fig3:ml2} indicate negative samples which will be separated from the matched feature space. A multi-classification method may obtain a weaker representation for a specific kin-type because of the large difference of inherited features among different kin-type images. A joint learning method has the advantage of the generalization of multi-class training and the representation of individual verification models. Hence, identification methods based on joint learning not only repulse negative pairs of different kinship types but also push the potential negative images to the target feature space, which is illustrated in Fig.~\ref{fig3:ml3}.

\subsubsection{Real World Kinship Distribution and Dataset Bias}
Note that the proportion of positive and negative samples is highly unbalanced for existing kinship verification datasets. This unbalanced distribution has a negative impact on different applications. Take the online
family picture organization application for example. The problem is to determine the matched pairs of images for a specific kinship relationship when the number of kin-related samples only contains a small portion of the entire dataset. Another example is that, when searching for missing children, to retrieve a picture that looks the most like the son of the parents in which the majority of these samples are negative samples.
    
\section{Joint Learning of Kinship Identification and Kinship Verification}
    \begin{figure}[tbp]
        \centering
        \includegraphics[width= 1\textwidth]{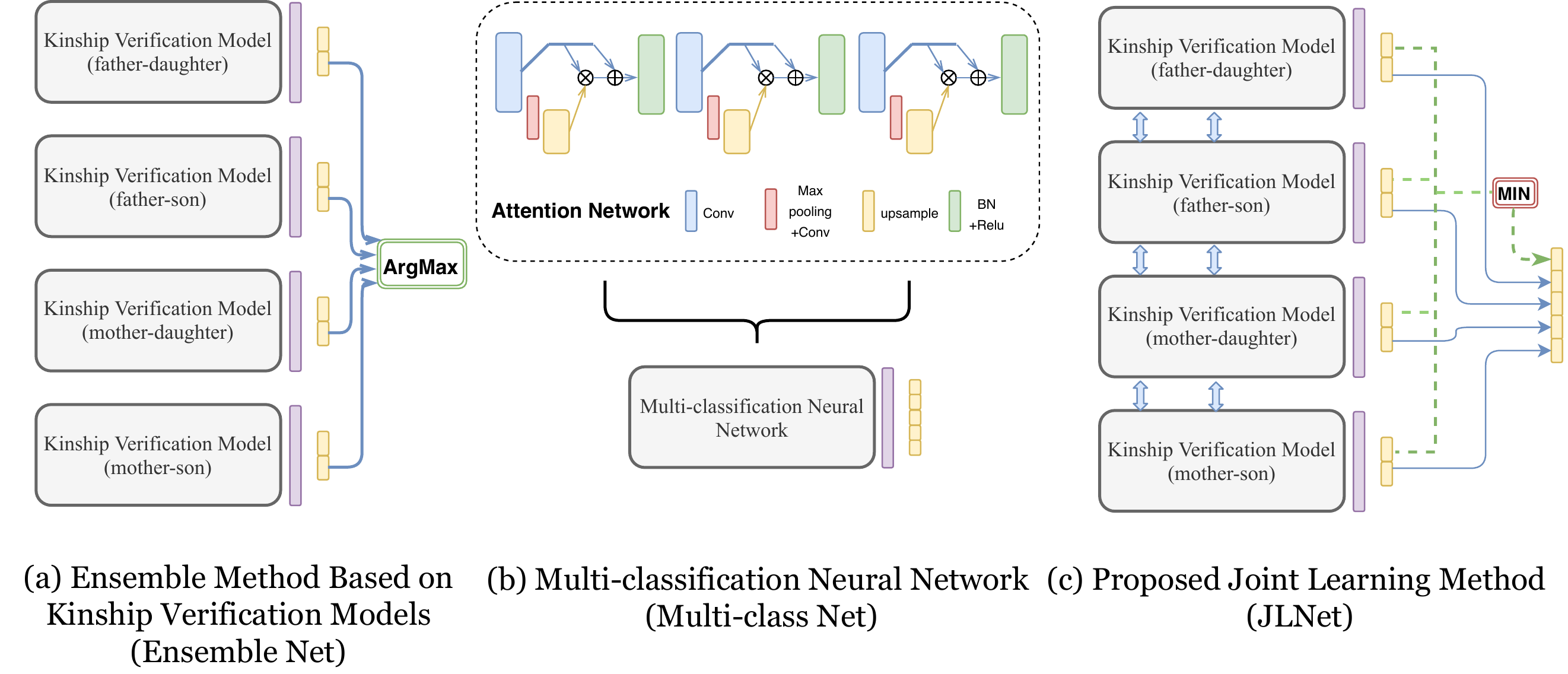}
        \caption{Structure of the approaches using four relationships as an example.}
        \label{fig:structure}
    \end{figure}
We propose a joint learning network (JLNet) based on the learning strategy shown in Fig. 4c aiming to utilize the representation capability of kinship verification models as well as making use of the advantages of multi classification. This approach consists of two major steps: the combination of different types of images and joint learning.

The main ideas of the approach are summarized as follows:
    \begin{enumerate}
        \item We utilize all different kin-types of image pairs to train each kinship model, not based on a specific type.
        \item Different models are trained jointly to differentiate
 negative kinship feature pairs from the matched model and to merge positive pairs as much as possible.
    \end{enumerate}
    
    Note that naively using a single classification network (Fig.~\ref{fig:structure}.a) or naively combining multiple verification networks (Fig.~\ref{fig:structure}.b) are not suitable approaches. As described above, our network (Fig.~\ref{fig:structure}.c) utilizes the advantage of both tasks. Without loss of generality, we outline our approach for four relationships: father-daughter (F-D), father-son (F-S), mother-daughter (M-D), mother-son (M-S).  
    

        
        
    

\subsection{Architecture of the Proposed Joint Learning Network (JLNet)}

    \begin{figure}[tbp]
    \centering
    \includegraphics[width= 1\textwidth]{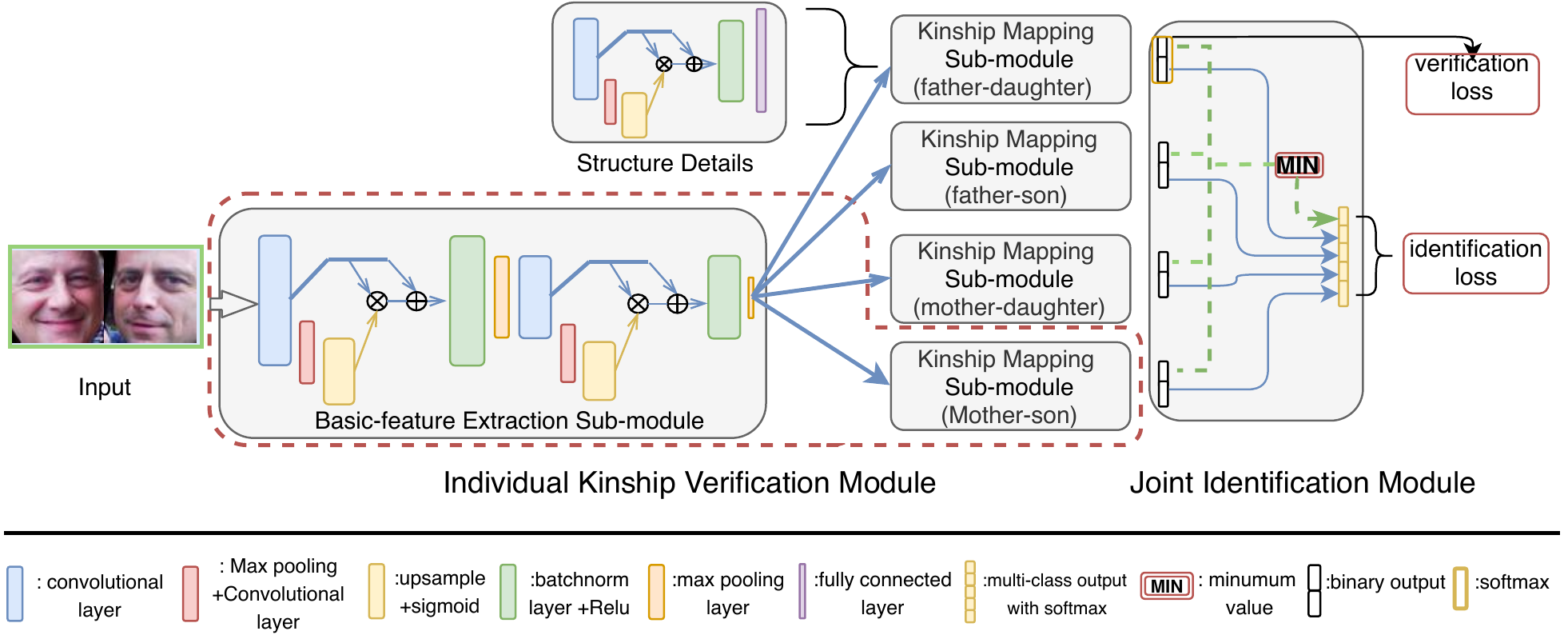}
    \caption{Architecture of our Joint Learning Network (JLNet)}
    \label{fig:JLNet}
    \end{figure}
    
The new Joint Learning Network(JLNet) is illustrated in Fig.~\ref{fig:JLNet}. The structure of JLnet consists of two parts: the individual Verification Module and the Joint Identification Module.

\subsubsection{Individual Kinship Verification Module}
As shown in Fig.~\ref{fig:structure}.c, each Individual Kinship Verification Module is defined as a binary classification problem. Let $\bm{S} = \left\{\left(I^\alpha_{p_i}, I^\alpha_{c_i}\right),i=1,2, \ldots,N, j=1,2, \ldots,N, \alpha= 1,2,3,4, \beta=1,2,3,4 \right\}$ be the training set of N pairs of images. And $\alpha\in\{1,2,3,4\}$ and $\beta\in\{1,2,3,4\}$ correspond to the following kinship types: father-daughter, father-son, mother-daughter, mother-son respectively. Then, the Individual Verification Module is defined by:

    \begin{equation}
        \hat{y}=\mathcal{D}_{\theta}^{n}\left(I_{p_i}^\alpha,I_{c_j}^\beta\right),
        \label{eq:binary}
    \end{equation}
where $I_{p_i}^\alpha \in \mathbb{R}^{H \times W \times 3}$ $i$th parent image from $\alpha$ type data set and $I_{c_j}^\beta \in \mathbb{R}^{H \times W \times 3}$ is the $j$th child image from $\beta$ type data set. The output $\hat{y}$ of each Individual kinship verification Module is a $1\times2$ vector. An Attention Network\cite{attenNet} is used as the basic architecture for each Individual Kinship Verification Module. As shown in Fig.~\ref{fig:JLNet}, the Attention Network uses a bottom-up top-down structure and consists of three attention stages. Each stage consists of one attention module and one residual structure. To exploit the shared information between the complimentary tasks, the parameters of the two stages of the Attention Network are shared to learn low-level and mid-level features from the input images. This forms the Basic-feature Extraction Sub-module. This Basic-feature Extraction Sub-module extracts the basic, generic facial features. Then, high-level features are extracted: four separate branches are added after the last layer (a max pool layer) of the Basic-feature Extraction Sub-module. Each branch focuses on one specific kin-type separately, resulting in four Kinship Mapping Sub-modules. Each of this sub-Module obtains the third stage of the Attention Network and focuses on different kinship types. 


\subsubsection{Joint Identification Module}
The binary outputs of each Individual Kinship Verification Module are ensembled. The binary output is described in Eq.~\ref{eq:binary}. The multiple output $\hat{O}$ of the kinship identification module is defined by:
    \begin{equation}
         \hat{O}_m=\left\{\begin{array}{ll}
        {\min _{n \in {\{1,2,3,4}\}} \mathcal{D}^n_{\theta}\left(I_{p_i}^\alpha,I_{c_j}^\beta\right)_{z=1},} & {\text { if } m=0} \\
        {\mathcal{D}^m_{\theta}\left(I_{p_i}^\alpha,I_{c_j}^\beta\right)_{z=2}} & {\text { if } m \neq 0} 
        \end{array}\right.,
    \label{eq:multiple}
    \end{equation}
where $m\in \{1,2,3,4,5\}$ represents the $m$th item of vector $\hat{O}$ and $z$ represents $z$th item of the output vector of $\mathcal{D}^n_{\theta}$. The output class $C$ is defined by:
    \begin{equation}
    C = \mathop{\arg\max}_{z\in\{1,2,3,4,5\}}\sigma(\hat{O})_{z},
    \label{eq:softmax}
    \end{equation}
where $\sigma(\cdot)$ is the softmax function.

During the training, the Weighted Cross Entropy loss is used for both kinship  verification and identification:
\begin{equation}
    \mathcal{L} = -\sum_{i=1}^{n}w_{n}log(\sigma(\cdot)_{n}),
    \label{eq:weighted_cross_entropy}
\end{equation}
where $n$ is the class label of the kinship verification or identification output and $\sigma(\cdot)_{n}$ is the $n$th output of the softmax function. The $w_{n}$ is weight of the $n$th class. The loss of the joint learning model is given by a weighted summation of the kinship verification loss (from binary outputs) and the kinship identification loss (from multiple outputs):
   \begin{equation}
        \mathcal{L} = \sum_{i=1}^{4}\lambda_{i}\mathcal{L}_{kv{i}}+\lambda_5\mathcal{L}_{kI},
    \end{equation}
where $\mathcal{L}_{kI}$ is the Weighted Cross Entropy loss of the kinship identification output given by Eq.~\ref{eq:weighted_cross_entropy} and $\lambda_i$ is the $i$th weight of each loss. 

\subsection{Comparative Methods}
\subsubsection{Ensemble Method based on Kinship Verification Models (Ensemble Net)}
Fig.~\ref{fig:structure}.a shows the structure of the Ensemble Method based on Kinship Verification Models (Ensemble Net). The Individual Kinship Verification Modules of the Ensemble Net have the same structure as JLNet. While testing, the Ensemble Net feeds the images into four kinship verification models simultaneously and ensembles four binary outputs. The output class $C$ is defined by:
\begin{equation}
        \text { C }=\left\{\begin{array}{ll}
        {0} & {\text { if }
        \operatorname*{max}_{n}\sigma(\mathcal{D}^n_{\theta}\left(I_{p_i}^\alpha,I_{c_j}^\beta\right))_{z=2}
         <0.5} \\
        {\operatorname*{argmax}_{n}\sigma(\mathcal{D}^n_{\theta}\left(I_{p_i}^\alpha,I_{c_j}^\beta\right))_{z=2},} & {\text{ otherwise }}
        \end{array}\right.,
    \label{eq:6}
\end{equation}
where $I^\alpha_{p_i}$ is the $i$th parent image from $\alpha$ type data set and $I^\beta_{c_j}$ is the $j$th child image from $\beta$ type data set.

\subsubsection{Multi-Classification Neural Network (Multi-class Net)}
The structure of the Multi-Classification Neural Network (Multi-class Net) is shown in Fig.~\ref{fig:structure}.b. Similar to the Ensemble Net, Multi-class has the same backbone with the Individual Kinship Verification Module of JLNet. The Multi-class Net handles the kinship identification task as a multiple classification problem:
 \begin{equation}
        \hat{y}=\mathcal{D}_{\theta}\left(I_{p_i}^\alpha,I_{c_i}^\beta\right),
\end{equation}
where $\bm{S} = \left\{\left(I^\alpha_{p_i}, I^\beta_{c_i}\right), i=1,2, \ldots,N, \alpha= 1,2,3,4,  \beta = 1,2,3,4\right\}$ and the output $\hat{y}$ is a $1\times5$ vector.

\section{Experiments}
   
\subsection{Unbias Dataset for Training and Testing}
    Three types of benchmark datasets are generated from the KinfaceWI and KinfaceWII datasets \cite{kinwild,kinwild2} consisting of four kinship types: father-daughter (F-D), father-son (F-S), mother-daughter (M-D), mother-son (M-S). To conduct the experiment on unbiased datasets, we re-balance the KinfaceWI and KinfaceWII datasets into three different benchmark datasets as follows:

    
    \begin{enumerate}
        \item \textit{Independent Kin-type Image Set}: This dataset has four independent subsets, where each subset contains one specific kinship type. This dataset simulates a dataset obtained by an ideal kinship classifier. The split of this image set is the same as KinfaceWI or KinfaceWII. The positive samples are the parent-children pairs with the same type of kinship. The negative samples are the pairs of unrelated parents and children within the same kin-type distribution. The positive and negative ratio is $1:1$.

        \item \textit{Mixed Kin-Type Image Set}: This dataset combines four different kin-type images taken from the KinfaceWI or KinfaceWII datasets resulting in the type ratio (father-daughter: father-son: mother-daughter: mother-son: negative pairs) to be $1:1:1:1:4$. This image set is used for both training and testing. Image pairs with kinship relations are denoted as positive samples. Negative samples are random image pairs without kinship relation but within the same type of distribution.
        \item \textit{Real-Scenario Kin-Type Image Set}: This dataset simulates the data distribution for real-world scenarios (e.g. retrieval of missing children). All the images in the kinfaceWI or KinfaceWII datasets are paired one by one, which leads to a highly unbalanced positive-negative rate. Taking KinfaceWII as an example, in each cross-validation, there will be 400 images (200 positive pairs) to be tested. All these images are paired one by one. The ratio of positive and negative pairs is $1:398$.
    \end{enumerate}

\subsection{Experimental Design}

All methods are trained on the Mixed Kin-Type Image Set. The dataset is divided into 5-folds and verified by a 5-cross validation. We use the same data augmentation for all methods. The data is augmented by randomly changing the brightness, contrast, and saturation of the image. Random grayscale variations, horizontal flipping, perspective changes, and resizing and cropping are also included. All images have the same size \(64\times 64\times 3\), and the batch size is set to be 64. 

\subsubsection{Proposed Joint Learning Method (JLNet)}
The training scheme of JLNet is divided into two phases. The first one is to train the network parameters for the four models independently. The weighted cross entropy is used for updating and the weight list is set to be $[0.25,8]$ for each verification output. The second phase is to update network parameters jointly by using both binary and multiple-outputs. The weight matrix of the cross-entropy of the multiple outputs is set to $[0.18,2,2,2,2]$, and $\lambda_i$ of the total loss is $1:1:1:1:10$ respectively. Adam is used as optimizer and the learning rate is set to $10^{-4}$. Since there is no public code available for the attention network, we re-implemented the attention network from scratch. During testing of the kinship verification of each individual kin-type, the binary output of the matched Individual Kinship Verification Module is taken as the final result. During testing of the kinship identification task, both the binary outputs (for kinship verification) and multiple outputs (for kinship identification) are used. A combined result based on the confidence of these two types of outputs are taken as the final result.
\paragraph{Ablation Study}
\begin{itemize}
    \item \textit{Joint Learning without Backpropagation of Multiple Outputs (JLNet$^\dag$)}: To assess the performance of additional multi-classification outputs, the structure of JLNet$^\dag$ is kept the same as JLNet. Further, JLNet$^\dag$ is trained in the same way as JLNet, but without using multiple output results for parameter updating.
    \item \textit{Joint Learning using Multiple Outputs for Kinship Identification (JLNet$^\ddag$)}: We use the trained model of JLNet directly but only the multiple output is taken as the final result during testing.
\end{itemize}

\subsubsection{Experiments and Comparison}
\paragraph{Ensemble Net}
For Ensemble Net, we provide two ways to train the models:
        \begin{itemize}
            \item \textit{Ensemble Net*}: Each verification model is trained separately on the Independent Verification Image Set, which is the same as \cite{attenNet}. This means that each independent kinship verification module is only trained on matched data. 
            \item\textit{Ensemble Net}: Each verification model is trained on the Mixed-Type Image Set, which is the same as the training data of JLNet and Multi-class Net. Adam is used and the learning rate was set to be $10^{-4}$. The weights of the cross entropy are $0.25,8$.
        \end{itemize}
\paragraph{Multi-Class Net}
Also for the Multi-Class Net, Adam is used as an optimizer. The learning rate is again $10^{-4}$. A weight list of [0.1,1,1,1] is used for the weighted Cross Entropy loss.

\subsection{Results \& Evaluation}
The methods are evaluated on the different datasets. Five-cross validation is used as the evaluation protocol. As a reminder, Ensemble Net* is trained on the Independent Kin-Type Kinship Image Set, JLNet$^\dag$ is trained without Backpropagation of Multiple Outputs, and JLNet$^\ddag$ uses multiple outputs as the final result. The results are shown in Table \ref{tb:acc_set1}-\ref{tb:F10_set3_f2}.
\subsubsection{Results for the Independent Kin-Type Image Set}
    \begin{table}[]
        \centering
        \caption{The accuracy of different methods through 5-fold cross-validation on the Independent Kin-Type Image Set.}
        \begin{adjustbox}{max width=1\textwidth}
                \begin{tabular}{lcccccccccc}
                \hline
                \multicolumn{1}{c}{}  & \multicolumn{5}{c}{\textbf{KinfaceWI}}  & \multicolumn{5}{c}{\textbf{KinfaceWII}}   \\ \cline{2-11} 
                \multicolumn{1}{c}{\multirow{-2}{*}{\textbf{Methods}}} & \textit{F-D} & \textit{F-S}              & \textit{M-D}     & \textit{M-S}           & \textit{Mean}                                      & \textit{F-D}         & \textit{F-S}      & \textit{M-D}                                      & \textit{M-S}     & \textit{Mean}     \\ \hline
                Ensemble Net*  & 0.7017    & 0.7506      & 0.7410     & 0.615 & \textbf{0.7021}    & 0.746                     & 0.7440         & 0.7520        & 0.7320            & \textbf{0.7435}        \\ \hline
                Multi-class Net      & 0.6463       & 0.6797       & 0.6650        & 0.5770     & 0.6420            & 0.5880       & 0.6240        & 0.6200     & 0.5920     & 0.6060     \\
                Ensemble Net     & 0.6425       & 0.6321     & 0.6382    & 0.577   & 0.6224    & 0.6060          & 0.6000     & 0.5860    & 0.6260      & 0.6045       \\

                JLNet$^\dag$  & 0.6534  & 0.6991   & 0.6539    & 0.5772     & 0.6459        & 0.6160    & 0.6100    & 0.600    & 0.6500      & 0.6190    \\
        
                JLNet    & 0.6608 & 0.7309 & 0.7207 & 0.5897 & \textbf{0.6755} & 0.6800 & 0.7140 & 0.6860 & 0.7060 & \textbf{0.6965} \\ \hline
                \end{tabular}
        \end{adjustbox}
    \label{tb:acc_set1}
    \end{table}

    \begin{table}[]
        \caption{F1 scores of different methods through 5-fold cross-validation on Independent Kin-Type Image Set}
            \begin{adjustbox}{max width=1\textwidth}
                \begin{tabular}{lccccclllll}
                \hline
         & \multicolumn{5}{c}{\textbf{KinfaceWI}}  & \multicolumn{5}{c}{\textbf{KinfaceWII}}         \\ \cline{2-11} 
                \multirow{-2}{*}{\textbf{Methods}} & \textit{F-D}                                      & \textit{F-S}    & \textit{M-D}          & \textit{M-S}     & \textit{Mean}   & \multicolumn{1}{c}{\textit{F-D}} & \multicolumn{1}{c}{\textit{F-S}} & \multicolumn{1}{c}{\textit{M-D}} & \multicolumn{1}{c}{\textit{M-S}} & \multicolumn{1}{c}{\textit{Mean}} \\ \hline
                Ensemble Net*                      & 0.6915                                            & 0.7472                                             & 0.7566                                             & 0.6648                                             & \textbf{0.7150}                                              & 0.7671                           & 0.7589                           & 0.7690                            & 0.7607                           & \textbf{0.7639}                   \\ \hline
                Multi-class Net                    & 0.6084                                            & 0.6563                                             & 0.6767                                             & 0.5766                                             & 0.6295                                             & 0.5629                           & 0.6000                           & 0.6143                           & 0.5062                           & 0.5709                            \\
                Ensemble Net                       & 0.6639                                            & 0.6737                                             & 0.6735                                             & 0.6083                                             & \textbf{0.6548}                                             & 0.6213                           & 0.6439                           & 0.6051                           & 0.6399                           & 0.6276                            \\

                JLNet$^\dag$                       & 0.6301                                            & 0.6952                                             & 0.6496                                             & 0.5816                                             & 0.6391                                             & 0.6396                           & 0.6166                           & 0.6061                           & 0.6191                           & 0.6203                            \\

                JLNet                         & 0.6320 & 0.7087 & 0.7052 & 0.5657 & 0.6529 & 0.6585                           & 0.7211                           & 0.6939                           & 0.6847                           & \textbf{0.6896}                            \\ \hline
                \end{tabular}
        \end{adjustbox}
        \label{tb:F1_set1}
    \end{table}
    Table \ref{tb:acc_set1} shows the verification results for the different methods based on the Independent Kin-Type Kinship Image Set. For this image set, accuracy and F1 scores are used to evaluate the performance of kinship verification. All methods are trained on the Mixed Kin-type Image Set except for ensemble Net*. The results show that when trained on the same dataset,  JLNet outperforms all other approaches. When tested on the KinfaceWII dataset, JLNet outperforms Multi-Class Net with 9\% and Ensemble Net by  9.2\% on average accuracy. Considering the F1 score, JLNet outperforms Multi-Class Net with 11.9\% and Ensemble Net with 6.2\% on average. When comparing JLNet\dag and JLNet, it is shown that additional multi-outputs improve the results of the ensembled models. When compared with Ensemble Net, the accuracy of JLNet is lower than Ensemble Net. One of the reason is that each of the verification module of Ensemble Net is trained on one specific dataset. This may result in overfitting. JLNet provides better generalization than Ensemble Net*, as shown in the next session.
    
\subsubsection{Results on Mixed Kin-Type Kinship Image Sets}
    \begin{table}[tbp]
    \centering
    \caption{Macro F1 score and accuracy of kinship identification for the Mixed Kin-Type Kinship Image Set}
    \begin{adjustbox}{max width=1\textwidth}
        \begin{tabular}{lcccc}
                \hline
                \multicolumn{1}{c}{}                                   & \multicolumn{2}{c}{\textbf{KinfaceWI}} & \multicolumn{2}{c}{\textbf{KinfaceWII}} \\ \cline{2-5} 
                \multicolumn{1}{c}{\multirow{-2}{*}{\textbf{Methods}}} & \textit{macro F1}   & \textit{Accuracy}     & \textit{macro F1}   & \textit{Accuracy}      \\ \hline
                Ensemble Net*                                          & 0.3240              & 0.3723           & 0.2846              & 0.3319            \\ \hline
                Multi-class Net                                        & 0.5291              & 0.5494           & 0.4861              & 0.5225            \\
                Ensemble Net                                           & 0.4837              & 0.4887           & 0.4464              & 0.4564            \\

                JLNet$^\dag$                                          & 0.5155     & 0.5487  & 0.4648     & 0.4875  \\
                
                JLNet$^\ddag$                                            & \textbf{0.5507}     & 0.5880          & 0.5285   & 0.5535   \\
                
                JLNet(full)                                   & 0.5506    & \textbf{0.5993}          & \textbf{0.5343}    & \textbf{0.5790}    \\ \hline
                \end{tabular}
        \end{adjustbox}
        \label{tb:macroF1_acc_set2}
    \end{table}

\begin{figure}[tbp]
        \centering

        \begin{subfigure}[t]{0.31\textwidth}
            \includegraphics[width=\textwidth]{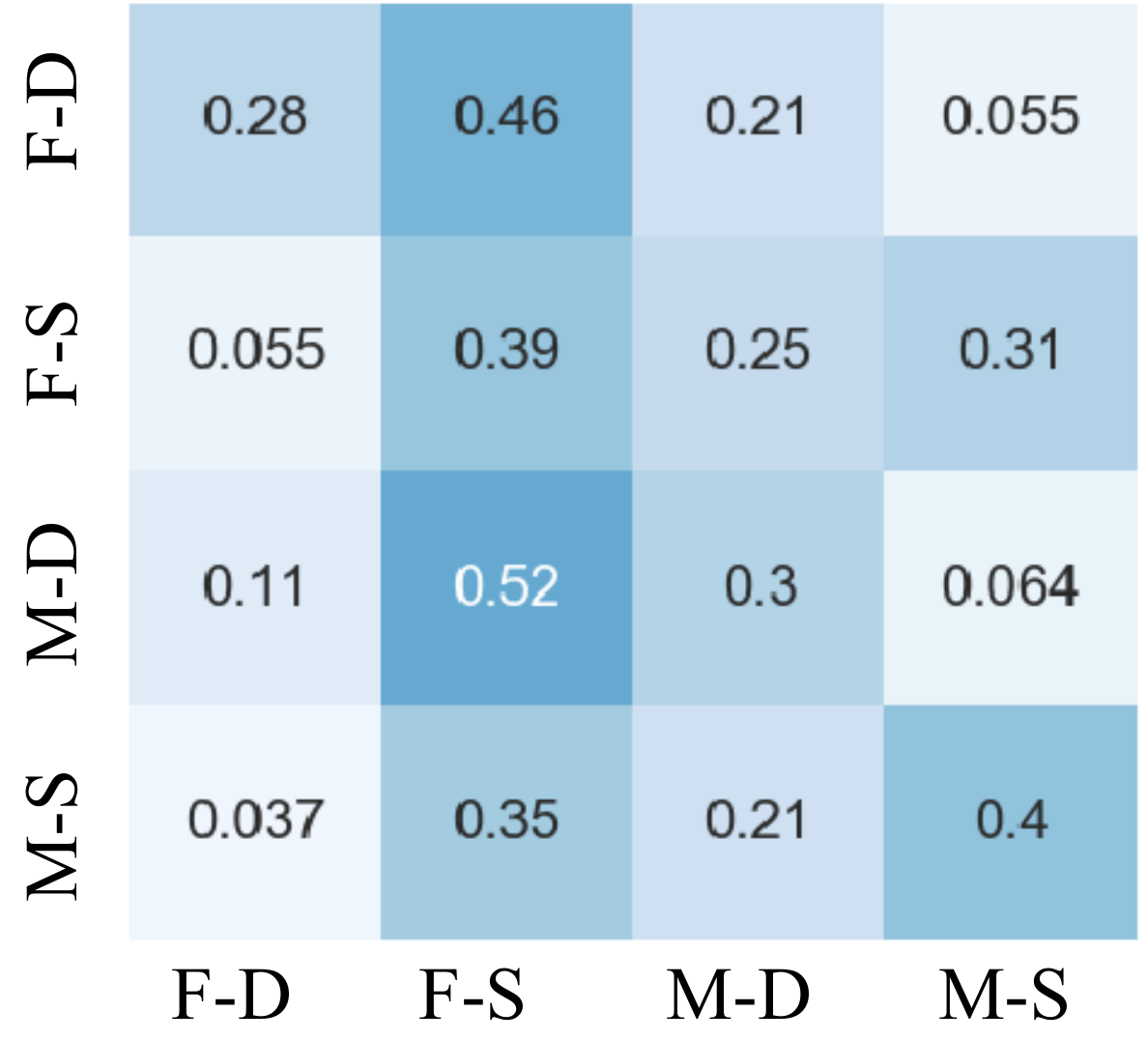}
            \caption{Ensemble Net*}
        \end{subfigure}
        ~
        \begin{subfigure}[t]{0.31\textwidth}
            \includegraphics[width=\textwidth]{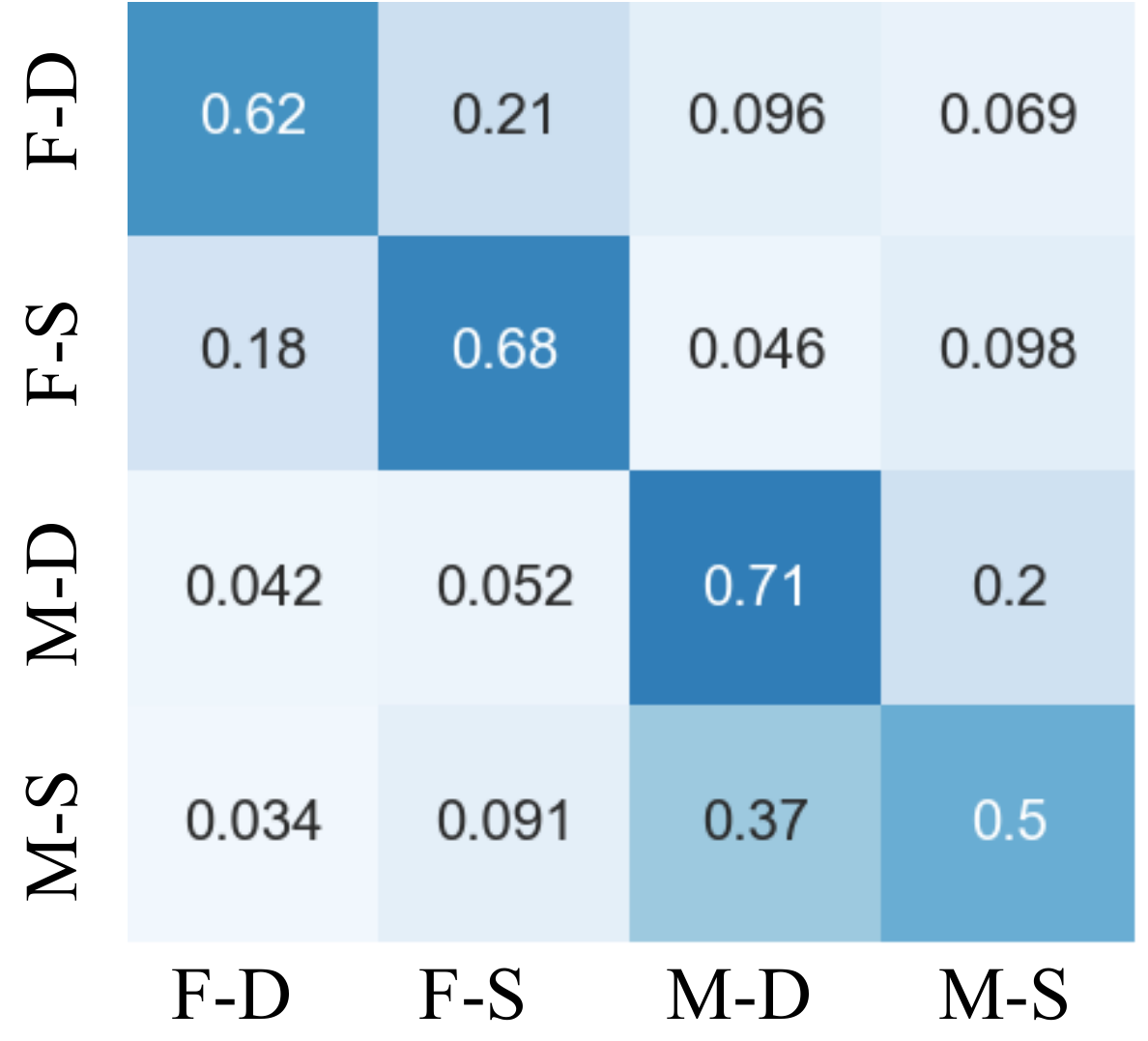}
            \caption{Multi-class Net}
        \end{subfigure}
        ~ 
         \begin{subfigure}[t]{0.31\textwidth}
            \includegraphics[width=\textwidth]{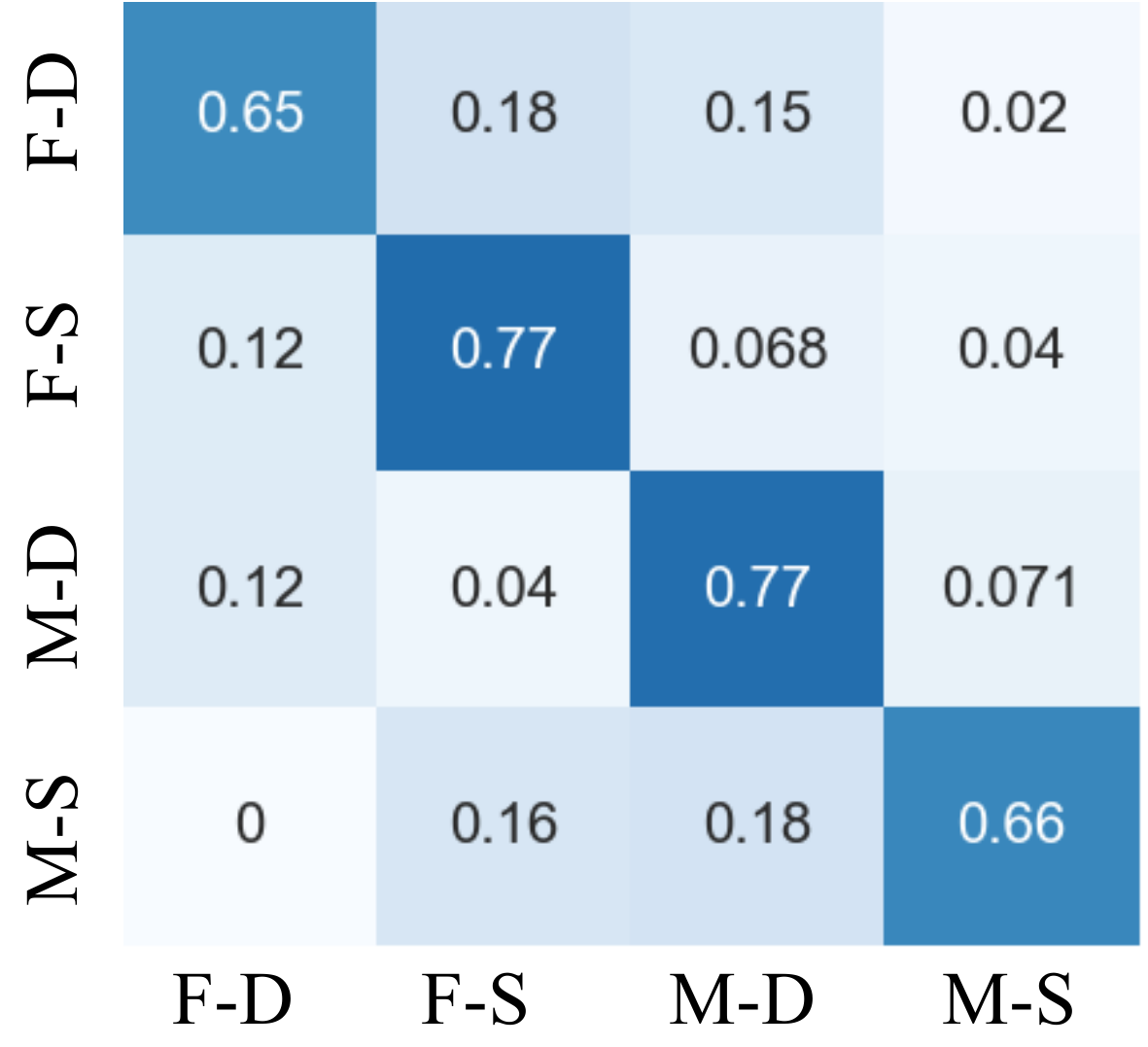}
            \caption{JLNet(full)}
        \end{subfigure}
        \caption{Confusion matrix for different experiments on the Mixed Kin-Type Image Set using the KinfaceWI dataset. Negative samples are excluded)}
        \label{fig:CM_kfw1_attenNet}
    \end{figure}

  Table \ref{tb:macroF1_acc_set2} shows the results of macro F1 scores and accuracy for the kinship identification task using the Mixed Kin-Type Kinship Image Set. The results show that the performances of JLNet outperforms the ensemble and multi-class net methods. Moreover, macro F1 scores show that JLNet(full) outperforms Ensemble Net* with 22.7\% on KinfaceWI and with 25.0\% on KinfaceWII. Moreover, JLNet(full) outperforms Ensemble Net* with 22.7\% on KinfaceWI and 24.7\% on KinfaceWII.  As shown in Fig.~\ref{fig:CM_kfw1_attenNet}, Ensemble Net* may lead to indecisive results. The independently trained verification models can lead to overfitting and results in weak generalization capabilities. JLNet obtained the highest performance. In Fig.~\ref{fig:CM_kfw1_attenNet}, it is shown that the joint learning method provides indecisive results. To this end, the joint learning method JLNet(full) obtains the best performance for kinship identification on the Mixed Kin-type Kinship Image Set.
\subsubsection{Results on Real Scenario Sample Set}
    \begin{table}[tb]
    \centering
        \caption{F10 score and accuracy for different methods on the Real-Scenario Set using KinfaceWI dataset. F10(all) represents the average of F10 scores for all different labels (the negative label is also included)}
            \begin{adjustbox}{max width=1\textwidth}
                \begin{tabular}{lccccccc}
                \hline
                \multicolumn{1}{c}{}                                   & \multicolumn{7}{c}{\textbf{KinfaceWI}}                                                                                                                                                                                                                                                                                                                                                                 \\ \cline{2-8} 
                \multicolumn{1}{c}{\multirow{-2}{*}{\textbf{methods}}} & \textit{F-D}                                        & \textit{F-S}                                        & \textit{M-D}                                        & \textit{M-S}                                        & \textit{mean}                                               & \textit{F10(all)}                                            & \textit{Accuracy}                                       \\ \hline
                Ensemble Net*                                          & 0.0886                                              & 0.1179                                              & 0.1236                                              & 0.1003                                              & 0.1076                                                      & 0.1830                                                      & 0.4807                                             \\ \hline
                Multi-class Net                                        & 0.1548                                              & 0.2951                                              & 0.3047                                              & 0.1539                                              & 0.2271                                                      & 0.2947                                                      & 0.5618                                             \\
                Ensemble Net                                           & 0.1508                                              & 0.2791                                              & 0.2740                                              & 0.1378                                              & 0.2104                                                      & 0.2596                                                      & 0.4537                                             \\
               
               JLNet$^\dag$                                           & 0.1522                                              & 0.2966                                              & 0.2937                                              & 0.1569                                              & 0.2249                                            & 0.2985                                             & 0.5901                                             \\

                JLNet$^\ddag$                                             & \textbf{0.1742} & 0.3235 & 0.3123 & 0.1620& 0.2430 & 0.3287& 0.6681 \\
                 
                JLNet(full)                                  & 0.1715                                              & \textbf{0.3241}                                              & \textbf{0.3198}                                            & \textbf{0.1669}                                              & \textbf{0.2456}                                             & \textbf{0.3459}                                                      & \textbf{0.7439}                                             \\ \hline
                \end{tabular}
            \end{adjustbox}
    \label{tb:F10_set3_f1}
    \end{table}      
    
    \begin{figure}[tbp]
        \centering
        \begin{subfigure}[b]{0.31\textwidth}
            \includegraphics[width=\textwidth]{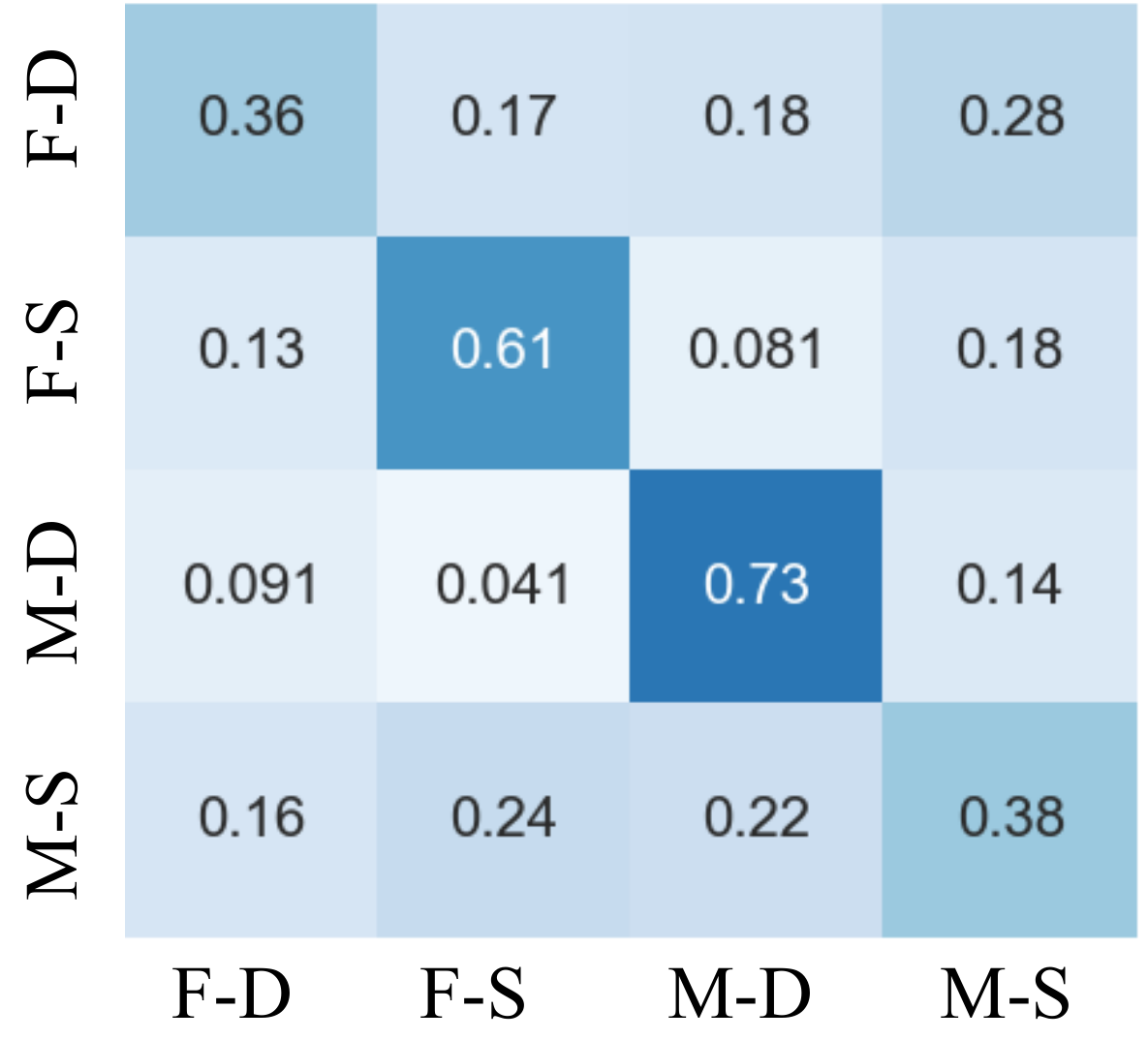}
            \caption{Multi-class Net}
        \end{subfigure}
        ~ 
        \begin{subfigure}[b]{0.31\textwidth}
            \includegraphics[width=\textwidth]{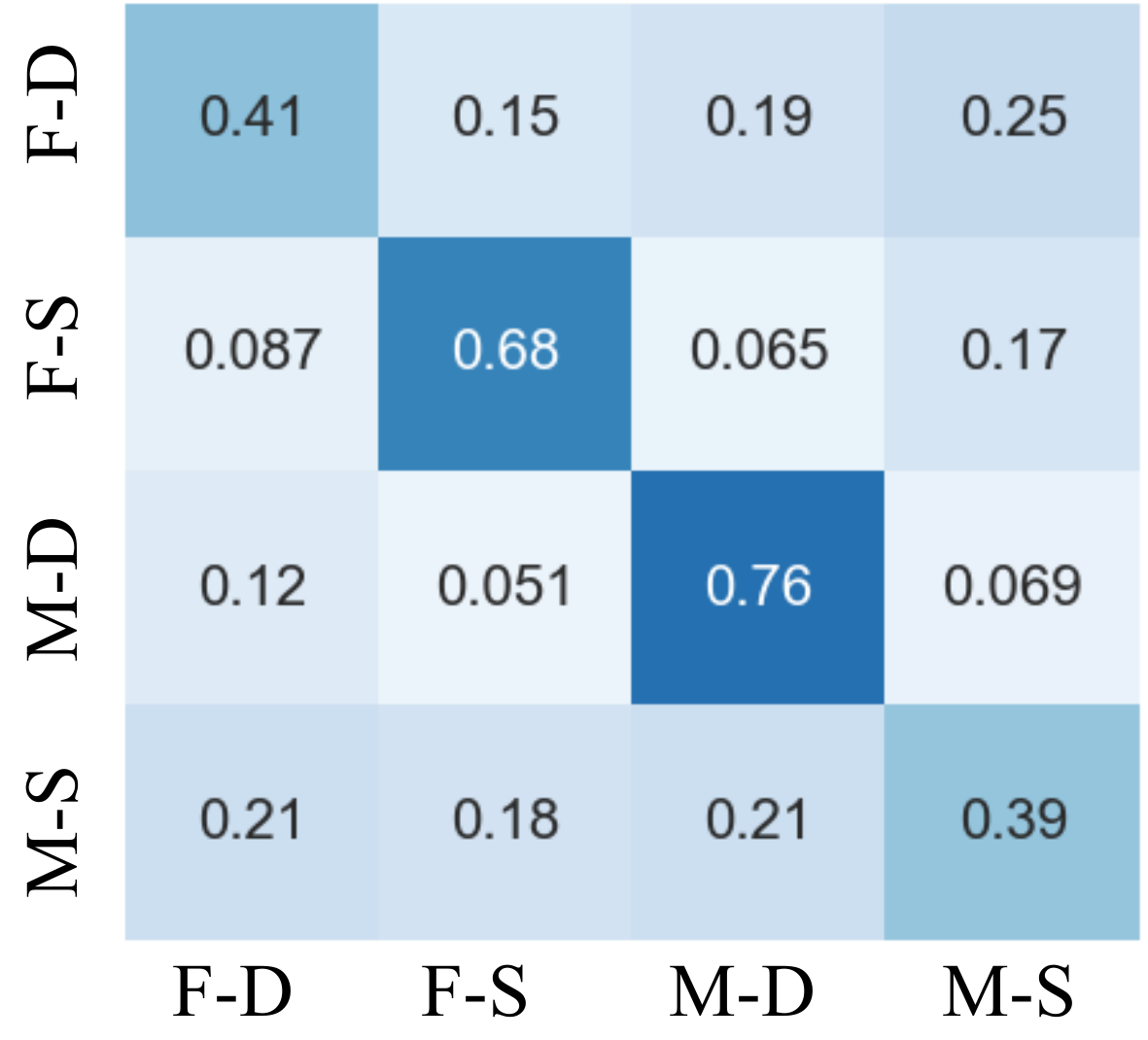}
            \caption{JLNet}
        \end{subfigure}
        ~
        \begin{subfigure}[b]{0.31\textwidth}
            \includegraphics[width=\textwidth]{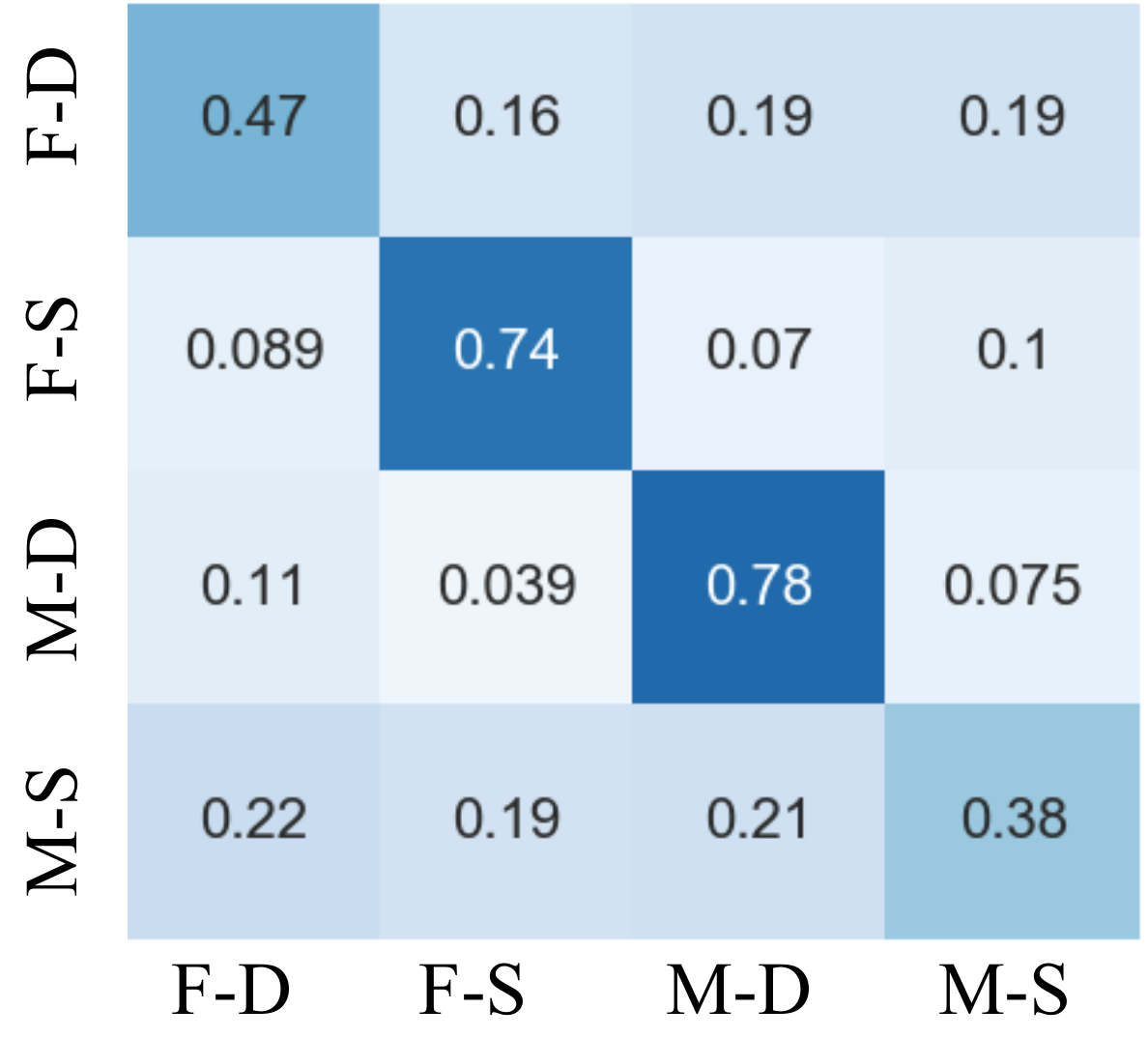}
            \caption{JNet(full)}
        \end{subfigure}
        \caption{Confusion matrix for different experiments on the Real-Senario Image set using the KinfaceWI dataset. Negative samples are excluded)}
        \label{fig:CM_kfw2_attenNet}
    \end{figure}
    
    \begin{table}[tb]
    \centering
        \caption{F10 score and accuracy for different methods on the Real-Scenario Set using KinfaceWII dataset. F10(all) represents the average of F10 scores for all different labels (the negative label is also included)}
        \begin{adjustbox}{max width=1\textwidth}
                      \begin{tabular}{lccccccc}
            \hline
            \multicolumn{1}{c}{}                                   & \multicolumn{7}{c}{\textbf{KinfaceWII}}                                                                                                                                                                                                                                                                                                                                                      \\ \cline{2-8} 
            \multicolumn{1}{c}{\multirow{-2}{*}{\textbf{methods}}} & \textit{F-D}                                       & \textit{F-S}                                        & \textit{M-D}                                       & \textit{M-S}                                        & \textit{mean}                                       & \textit{F10(all)}                                            & \textit{Accuracy}                                       \\ \hline
            Ensemble Net*                                          & 0.0469                                             & 0.0713                                              & 0.0726                                             & 0.0904                                              & 0.0703                                              & 0.1498                                                      & 0.4647                                             \\ \hline
            Multi-class Net                                        & 0.1468                                             & 0.1972                                              & 0.1853                                             & 0.1076                                              & 0.1592                                              & 0.2528                                                      & 0.6240                                             \\
            Ensemble Net                                           & 0.1399                                             & 0.1681                                              & 0.1496                                             & 0.0900                                              & 0.1369                                              & 0.2075                                                      & 0.4874                                             \\

             JLNet$^\dag$                                          & 0.1413                                             & 0.1757                                              & 0.1624                                             & 0.0962                                             & 0.1439                                             & 0.2303                                                      & 0.5730                                             \\

             JLNet$^\ddag$                                            & 0.1620 & 0.2133 & 0.2127 & 0.1225 & 0.1776 & 0.2735 & 0.6547 \\

            JLNet(full)                                 & \textbf{0.1867} & \textbf{0.2134} & \textbf{0.2296} & \textbf{0.1296}  & \textbf{0.1898}  & \textbf{0.3003}         & \textbf{0.7398} \\ \hline
            \end{tabular}
            \end{adjustbox}
        \label{tb:F10_set3_f2}
    \end{table}

    Tables \ref{tb:F10_set3_f1} and \ref{tb:F10_set3_f2} show the results of the F10 score and accuracy for the kinship identification task in a real-world scenario. We focus more on recall than precision, so the F10 score is used to emphasize on the recall rate. The results show that JLNet(full) obtains the best performance on both KinfaceWI-based Real-Scenario data and KinfaceWII-based Real-Scenario data. The results show that the  JLNet(full) outperforms all the other approaches for both KinfaceWI and KinfaceWII.  From the confusion matrix in Fig.~\ref{fig:CM_kfw2_attenNet}, it is interesting to note that father-son and mother-daughter relations are more distinguishable than other kin-types. We argue that the manifold of pairs with the same gender is easier to be learned.

\section{Conclusion}
     In this paper, we presented a new approach for kinship identification by joint learning. Experimental results show that joint learning with kinship verification and identification improves the performance of kinship identification. To our knowledge, this is the first approach to handle the kinship identification tasks by using deep neural networks jointly. Since this method is not restricted to any neural network, a better architecture can further improve the performance for kinship identification.

%
%
\bibliographystyle{splncs04}
\bibliography{main}
\end{document}